\documentclass{article}
\usepackage{spconf,amsmath,graphicx}
\usepackage{multirow}
\usepackage[table,xcdraw]{xcolor}

\title{Kwame: A Bilingual AI Teaching Assistant for Online SuaCode Courses}
\name{George Boateng}
\address{ETH Zurich, Switzerland}

\begin{document}
%
\maketitle
\begin{abstract}
Introductory hands-on courses such as our smartphone-based coding course, SuaCode require a lot of support for students to accomplish learning goals. Online environments make it even more difficult to get assistance especially more recently because of COVID-19. Given the multilingual context of SuaCode students  — learners across 42 African countries that are mostly Anglophone or Francophone — in this work, we developed a bilingual Artificial Intelligence (AI) Teaching Assistant (TA) — Kwame — that provides answers to students’ coding questions from SuaCode courses in English and French. Kwame is a Sentence-BERT (SBERT)-based question-answering (QA) system that we trained and evaluated offline using question-answer pairs created from the course’s quizzes, lesson notes and students’ questions in past cohorts. Kwame finds the paragraph most semantically similar to the question via cosine similarity. We compared the system with TF-IDF and Universal Sentence Encoder. Our results showed that fine-tuning on the course data and returning the top 3 and 5 answers improved the accuracy results. Kwame will make it easy for students to get quick and accurate answers to questions in SuaCode courses.
\end{abstract}
\begin{keywords}
Virtual Teaching Assistant, Question Answering, NLP, BERT, SBERT, Machine Learning, Deep Learning
\end{keywords}
\section{Introduction}
\begin{figure}[t]
  \centering
  \includegraphics[width=\linewidth]{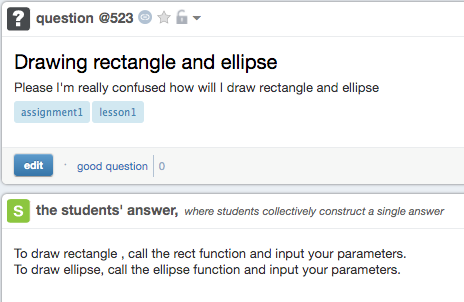}
  \caption{Course forum with question answering example}
  \label{fig:qa}
\end{figure}

Introductory hands-on courses such as our smartphone-based coding courses, SuaCode \cite{boateng2018,boateng2019,suacodeafrica} require a lot of support for students to accomplish learning goals (Figure \ref{fig:qa}). Offering assistance becomes even more challenging in an online course environment which has become important recently because of COVID-19 with students struggling to get answers to questions. Hence, offering quick and accurate answers could improve the learning experience of students. However, it is difficult to scale this support with humans when the class size is huge — hundreds of thousands — since students tend to ask questions whose answers could be found in course materials and also repetitively. 

There has been some work to develop virtual teaching assistants (TA) such as Jill Watson \cite{goel2016,goel2020}, Rexy \cite{benedetto2019}, and a physics course TA \cite{zylich2020}. All of these TAs have focused on logistical questions, and none have been developed and evaluated using coding courses in particular. Also, they have used one language (e.g. English). 

Given the multilingual context of our students  — learners across 42 African countries that are mostly Anglophone or Francophone — in this work, we developed a bilingual Artificial Intelligence (AI) TA — Kwame — that provides answers to students’ coding questions from our SuaCode courses in English and French. Kwame is named after Dr. Kwame Nkrumah the first President of Ghana and a Pan Africanist whose vision for a developed and empowered Africa resonates with the goals of our work. Kwame is a Sentence-BERT-based question-answering (QA) system that is trained using the SuaCode course material \cite{suacode} and evaluated offline using accuracy and time to provide answers. Kwame finds the paragraph most semantically similar to the question via cosine similarity. We compared Kwame with other approaches and performed a real-time implementation. Our work offers a unique solution for online learning, suited to the African context.

The rest of this paper is organized as follows: In Section 2, we talk about the background of SuaCode and related work. In Section 3, we describe Kwame's system architecture and real-time implementation. In Section 4, we describe our data and preprocessing approach. In Section 5, we describe our experiments and evaluation. In Section 6 we present and discuss the results. In Section 7, we describe the limitations of this work and future work. We conclude in Section 8.

\section{Background and Related Work}
In this section, we give the background of the SuaCode course and describe related work on virtual teaching assistants and the QA sentence selection task.

\subsection{SuaCode Course}
In 2017, we developed a smartphone-based coding course in Ghana, SuaCode, that enabled students to learn to code on their phones circumventing the need for computers \cite{boateng2018}. We launched an online version of SuaCode in 2018 \cite{boateng2019} and expanded it beyond Ghana to all Africans, dubbed SuaCode Africa in 2019 \cite{suacodeafrica}. For our most recent cohort that run from June 2020 — SuaCode Africa 2.0, the course was offered in both English and French \cite{suacodeafrica2}. Over 2,300 students across 69 countries, 42 of which are in Africa applied. We accepted and trained 740 students \cite{suacodeafrica2}. Our long term vision is to leverage smartphones to teach millions across Africa how to code. With our students needing a lot of assistance in their first coding course ever, we have thus far relied on human facilitators to provide support and answer students’ questions. For example, in SuaCode Africa 2.0, facilitators contributed over 1,000 hours of assistance time for an 8-week period and helped to achieve an average response time of 6 minutes through the course. This approach is however not scalable as the number of students applying to SuaCode is increasing exponentially. An AI teaching assistant that provides accurate and quick answers to students' questions would reduce the burden on human teaching assistants, and provide an opportunity to scale learning without reducing the quality of education. 

\subsection{Virtual Teaching Assistants}
The first work on virtual TAs was done by Professor Ashok Goel at the Georgia Institute of Technology in 2016. His team built Jill Watson, an IBM Watson-powered virtual TA to answer questions on course logistics in an online version of an Artificial Intelligence course for master’s students \cite{goel2016}. It used question-answer pairs from past course forums. Given a question, the system finds the closest related question and returns its answer if the confidence level is above a certain threshold. Since then, various versions of Jill Watson have been developed to perform various functions: Jill Watson Q \& A (revision of the original Jill Watson) now answers questions using the class' syllabi rather than QA pairs. Jill Watson SA gives personalized welcome messages to students when they introduce themselves in the course forum after joining and helps create online communities; Agent Smith aids to create course-specific Jills using the course' syllabus \cite{goel2020}. Jill’s revision now uses a 2-step process. The first step uses commercial machine learning classifiers such as Watson and AutoML to classify a sentence into general categories. Then the next step uses their own proprietary knowledge-based classifier to extract specific details and gives an answer from  Jill's knowledge base using an ontological representation of the class syllabi which they developed. Also, the responses pass through a personality module that makes it sound more human-like. Jill has answered thousands of questions, in one online and various blended classes with over 4,000 students over the years.

Similar work has been done by other researchers who built a platform, Rexy for creating virtual TAs for various courses also built on top of IBM Watson \cite{benedetto2019}. The authors described Rexy, an application they built which can be used to build virtual teaching assistants for different courses and also presented a preliminary evaluation of one such virtual teaching assistant in a user study. The system has a messaging component (such as Slack), an application layer for processing the request, a database for retrieving the answers, and an NLP component that is built on top of IBM Watson Assistant which is trained with question-answer pairs. They use intents (the task students want to be done) and entities (the context of the interaction). The intents (71 of those, e.g., exam date) are defined in Rexy but the entities have to be specified by instructors as they are course-specific. For each question, a confidence score is determined and if it is below a threshold, the question is sent to a human TA to answer and the answer is forwarded to the student. The authors implemented an application using Rexy and deployed it in an in-person course (about recommender systems) with 107 students. After the course, the authors reviewed the conversation log identified two areas of requests (1) course logistics (e.g., date and time of exams) and (2) course content (e.g. definitions and examples about lecture topics).

In the work by Zylich et al. \cite{zylich2020}, the authors developed a question-answering method based on neural networks whose goal was to answer logistical questions in an online introductory physics course. Their approach entailed the retrieval of relevant text paragraphs from course materials using TF-IDF and extracting the answer from it using an RNN to answer 172 logistical questions from a past physics online course forum. They also applied their system to answering 18 factual course questions using document retrieval without extracting an answer.

These works do not focus on answering questions about the course content but rather, course logistics such as the format for submitting assignments, goals of the course, etc. Our work will bridge that gap by focusing on providing answers to questions about course content like “how do I draw a circle at the center of my screen?”. Additionally, the technical details of some of the systems and their evaluations both offline and online are not available (e.g. \cite{goel2016,goel2020,benedetto2019}) and hence they are not that easy to build upon and compare with. Our work provides these details along with our systematic evaluation to allow other researchers to replicate our work and also build upon it to create similar systems for their context. Also, all the previous systems only work for one language (e.g. English) whereas our system works for English and French. Additionally, none of those TAs have been developed for coding courses in particular. Coding courses have specific vocabulary which become more important when answers to content questions have to be provided. They pose unique challenges that need to be addressed to have systems that work well. Our work addresses these.

\subsection{Question Answering: Answer Sentence Selection}
Within the domain of question answering, there are two paradigms: Machine reading and sentence selection. Machine reading entails selecting a span of text in a paragraph that directly answers the question. To extract the answer span, either the paragraph is provided (e.g., in the SQuAD question answering challenge \cite{rajpurkar2016}), or the relevant paragraph or set of paragraphs have to be retrieved from various documents first (e.g., as used in the physics QA system described in the previous section \cite{zylich2020}). TF-IDF is mostly used for the paragraph(s) or document retrieval \cite{chen2017}. For answer extraction, RNNs have been used \cite{zylich2020,chen2017} but BERT-based models are the current state-of-the-art \cite{devlin2018}.

The task of answer sentence selection entails predicting which sentence among a group of sentences correctly answers a question. Two examples of public challenges within this domain are the WikiQA and Natural Questions. The WikiQA dataset contains 3K questions from Bing queries with possible answers selected from Wikipedia pages \cite{yang2015}. Natural Question contains 307K questions based on Google searches \cite{kwiatkowski2019}. Various approaches have been used to tackle these challenges such as TF-IDF and unigram count, and  word2vec with CNN \cite{yang2015} and BERT and RoBERTa \cite{garg2019}. BERT based models are currently state-of-the-art for natural language processing tasks and they are used in our work.

In this work, we use the answer sentence selection paradigm as opposed to the machine reading one. We do so because for our context of answering questions in online courses, especially content questions, short answers are generally not adequate but rather, whole contexts that could span multiple continuous sentences. Hence, extracting an answer span is not necessary and could even be inadequate and more error-prone. This point is also made by Zylich et al. \cite{zylich2020} after assessing the misclassifications in their evaluation using machine reading to answer physics course questions. We as a result focus on selecting and providing whole paragraphs as answers.

\section{Kwame's System Architecture}
\begin{figure}[t]
  \centering
  \includegraphics[width=\linewidth]{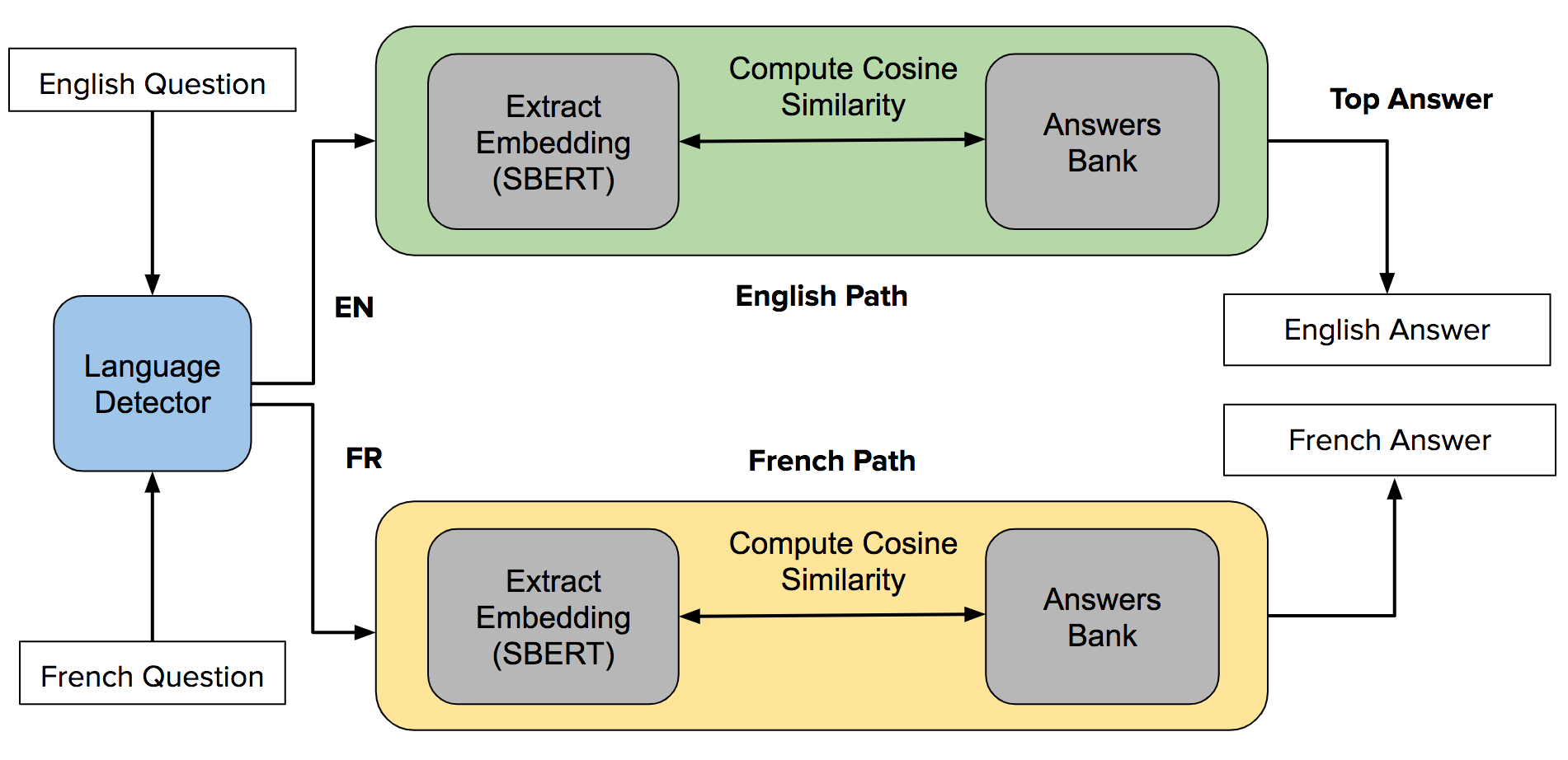}
  \caption{System Architecture of Kwame}
  \label{fig:Kwame-architecture}
\end{figure}

Kwame's system model is Sentence-BERT (SBERT), a modification of the BERT architecture with siamese and triplet network structures for generating sentence embeddings such that semantic similar sentences are close in vector space \cite{reimers2019} (Figure \ref{fig:Kwame-architecture}). The SBERT network was shown to outperform state-of-the-art sentence embedding methods such as BERT and Universal Sentence Encoder for semantic similarity tasks. We used the multilingual version \cite{reimers2020}. 

\begin{figure}[t]
  \centering
  \includegraphics[width=\linewidth]{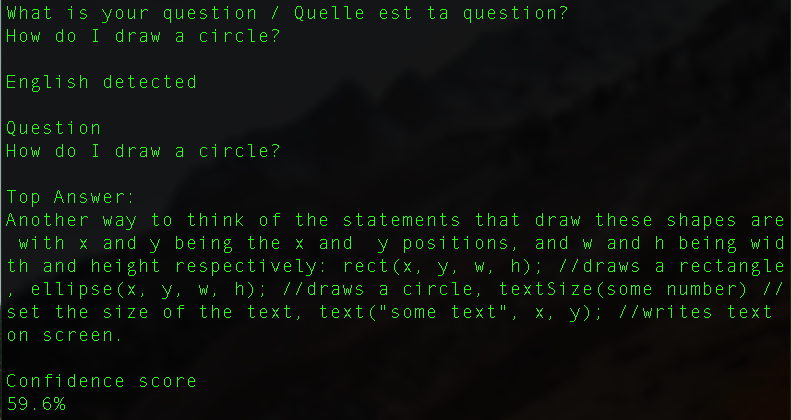}
  \caption{Example of a real-time implementation of Kwame (English)}
  \label{fig:Kwame}
\end{figure}

\begin{figure}[ht]
  \centering
  \includegraphics[width=\linewidth]{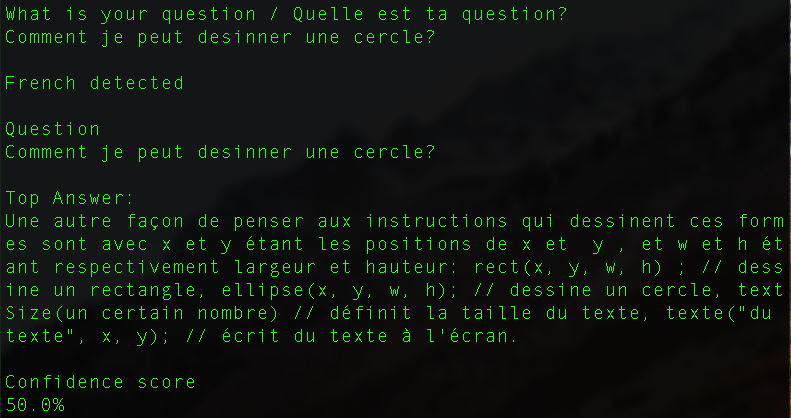}
  \caption{Example of a real-time implementation of Kwame (French)}
  \label{fig:Kwame-fr}
\end{figure}

We also created a basic real-time implementation of Kwame using the SBERT model fine-tuned with the course data. A user types a question, Kwame detects the language automatically (using a language detection library) and then computes cosine similarity scores with a bank of answers (described next) corresponding to that language, retrieves, and displays the top answer along with a confidence score which represents the similarity score (Figure \ref{fig:Kwame} and \ref{fig:Kwame-fr}).

\section{Dataset and Preprocessing}
We used the course materials from our “Introduction to Programming” SuaCode course written in English and French. Each document contains text organized by paragraphs that explain concepts along with code examples, tables, and figures which were removed during preprocessing. Also, the course has multiple choice quizzes for each lesson and the answer to each question has a corresponding paragraph in the course material. 

In this work, we used lesson 1, “Basic Concepts in the Processing Language” to create 2 types of question-answer pairs (1) quiz-based \textit{(n=20)} using the course’ quiz questions and (2) student-based \textit{(n=12)} using real-world questions from students in past cohorts along with the corresponding paragraph answers in the course materials. There were 39 paragraphs and hence a random baseline of  2.6\% for answer accuracy.

\begin{table*}[htb]
\centering
\caption{Accuracy and Duration Results}
\label{tab:results}
\resizebox{0.6\textwidth}{!}{%
\begin{tabular}{|l|c|c|c|c|c|c|}
\hline
\multicolumn{1}{|c|}{\textbf{Model}} & \multicolumn{4}{c|}{\textbf{Accuracy (\%)}} & \multicolumn{2}{c|}{\textbf{Duration  (secs per question)}} \\ \hline
\textbf{} & \multicolumn{2}{c|}{\textbf{English}} & \multicolumn{2}{c|}{\textbf{French}} & \multicolumn{1}{l|}{} & \multicolumn{1}{l|}{} \\ \cline{1-5}
\textbf{} & \textbf{Quiz} & \textbf{Student} & \textbf{Quiz} & \textbf{Student} & \multicolumn{1}{l|}{\multirow{-2}{*}{English}} & \multicolumn{1}{l|}{\multirow{-2}{*}{French}} \\ \hline
\rowcolor[HTML]{dcdcdc} 
TF-IDF (Baseline) & 30\% & 16.7\% & 45\% & 8.3\% & 0.03 & 0.02 \\ \hline
\rowcolor[HTML]{dcdcdc} 
Universal Sentence Encoder (USE) & 40\% & 25\% & 35\% & 16.7\% & 3.7 & 3.2 \\ \hline
\rowcolor[HTML]{C0C0C0} 
SBERT (regular) & 50\% & 25\% & 65\% & 8.3\% & 3.0 & 2.7 \\ \hline
\rowcolor[HTML]{C0C0C0} 
SBERT (trained) & 50\% & 25\% & 60\% & 8.3\% & 6.0 & 5.5 \\ \hline
\rowcolor[HTML]{C0C0C0} 
SBERT (fine-tuned with Quiz) & \textbf{65\%} & 16.7\% & \textbf{70\%} & 8.3\% & 6.0 & 6.0 \\ \hline
\rowcolor[HTML]{C0C0C0} 
SBERT (fine-tuned with Student) & 60\% & \textbf{58.3\%} & 65\% & \textbf{58.3\%} & 5.8 & 5.6 \\ \hline
\end{tabular}
}
\end{table*}

\begin{table*}[htb]
\centering
\caption{Top 1, 3, and 5 Accuracy Results for SBERT}
\label{tab:top_n_results}
\resizebox{0.9\textwidth}{!}{%
\begin{tabular}{|l|c|c|c|c|c|c|c|c|c|c|c|c|}
\hline
\rowcolor[HTML]{FFFFFF} 
\multicolumn{1}{|c|}{\cellcolor[HTML]{FFFFFF}\textbf{Model}} & \multicolumn{12}{c|}{\cellcolor[HTML]{FFFFFF}\textbf{Accuracy (\%)}} \\ \hline
\rowcolor[HTML]{FFFFFF} 
 & \multicolumn{6}{c|}{\cellcolor[HTML]{FFFFFF}\textbf{English}} & \multicolumn{6}{c|}{\cellcolor[HTML]{FFFFFF}\textbf{French}} \\ \hline
\rowcolor[HTML]{FFFFFF} 
 & \multicolumn{3}{c|}{\cellcolor[HTML]{FFFFFF}\textbf{Quiz}} & \multicolumn{3}{c|}{\cellcolor[HTML]{FFFFFF}\textbf{Student}} & \multicolumn{3}{c|}{\cellcolor[HTML]{FFFFFF}\textbf{Quiz}} & \multicolumn{3}{c|}{\cellcolor[HTML]{FFFFFF}\textbf{Student}} \\ \hline
\rowcolor[HTML]{FFFFFF} 
 & \textbf{Top 1} & \textbf{Top 3} & \textbf{Top 5} & \textbf{Top 1} & \textbf{Top 3} & \textbf{Top 5} & \textbf{Top 1} & \textbf{Top 3} & \textbf{Top 5} & \textbf{Top 1} & \textbf{Top 3} & \textbf{Top 5} \\ \hline
\rowcolor[HTML]{dcdcdc} 
SBERT (regular) & 50 & 75 & 75 & 25 & 50 & 75 & 65 & 75 & 75 & 8.3 & 50 & 75 \\ \hline
\rowcolor[HTML]{dcdcdc} 
SBERT (trained) & 50 & 75 & 75 & 25 & 33 & 91.7 & 60 & 75 & 75 & 8.3 & 58 & 75 \\ \hline
\rowcolor[HTML]{C0C0C0} 
SBERT (fine-tuned with Quiz) & \textbf{65} & 80 & 75 & 16.7 & 50 & 83.3 & \textbf{70} & 75 & 75 & 8.3 & 33 & 91.7 \\ \hline
\rowcolor[HTML]{C0C0C0} 
SBERT (fine-tuned with Student) & 60 & \textbf{80} & \textbf{85} & \textbf{58.3} & \textbf{83.3} & \textbf{100} & 65 & \textbf{80} & \textbf{80} & \textbf{58.3} & \textbf{91.7} & \textbf{91.7} \\ \hline
\end{tabular}%
}
\end{table*}

\section{Experiments and Evaluation}
We performed various experiments to evaluate the accuracy of our proposed models and the duration to provide answers. We used 3 models. The first model — SBERT (regular) — is the normal SBERT model which has already been fine-tuned on various semantic similarity tasks. Hence, it has no course-specific customization. 

For the second model — SBERT (trained) —, we trained the SBERT model using weakly-labeled triplet sentences from the course materials in a similar manner as Reimers et al. \cite{reimers2019} to learn the semantics of the course’ text. For each paragraph, we used each sentence as an anchor, the next sentence after it in that paragraph as a positive example, and a random sentence in a random paragraph in the document as a negative example. We created a train-test split (75\% : 25\%) and trained the model using the triplet objective in Reimers et al. \cite{reimers2019}. 

For the third model, we explored fine-tuning the SBERT model separately using the quiz QA pairs — SBERT (fine-tuned with Quiz) — and student QA pairs — SBERT (fine-tuned with Student) — using the same triplet objective to enable finding paragraphs that are semantically similar to questions. 

We compared these models with TF-IDF and bi-grams as the baseline and Universal Sentence Encoder as a competitive alternative. The models were evaluated separately with the quiz and student QA pairs. To evaluate, we extracted each question’s embedding and then computed the cosine similarity between the question’s embedding and all the possible answers’ embeddings, and returned the answer with the biggest similarity score. We then computed the accuracy of the predictions and the average duration per question. We precomputed and saved the embeddings to ensure the performance is quick. Evaluations were performed on a MacBook Pro with 2.9 GHz Dual-Core Intel Core i5 processor.

In addition to this top 1 accuracy evaluation, we computed and compared top 3, and 5 accuracy results for the SBERT models similar to Zylich et al. \cite{zylich2020}. To perform the evaluation, Kwame returns the top 3 or 5 answers and we check if the correct answer is any of those 3 or 5 answers. 

\section{Results and Discussion}
The duration and accuracy results are shown in Table \ref{tab:results}. The duration results show that TF-IDF is the fastest method, followed by SBERT regular, USE, SBERT trained, and SBERT fine-tuned taking the most time of 6 seconds which is not very long nonetheless, compared to the average response time of 6 minutes in the recent SuaCode course (Table \ref{tab:results}). These results are consistent with the result of  Rana \cite{rana2019} who found a similar trend.

For the quiz data, TF-IDF has the worst performance of 30\% and 45\% for English and French respectively (but better than the random baseline of 2.6\%) with USE and SBERT (Regular) having better performance. 

Our SBERT model that we trained using the weakly-labeled data from the course materials did not perform better than SBERT regular. This result might suggest that using weakly-labeled data with the triplet objective might not be adequate to improve the results. It could also be due to the small number of samples. Further exploration is needed. 

The SBERT models that were fine-tuned on the QA task had the highest accuracies as expected. Overall, the models performed better for the quiz data than the student data. This result is expected since the quiz’s questions were created based on specific paragraphs in the text which served as answers. The students’ questions however were from the real-world and did not have specific answers used to create the question. Also, the questions were noisy with various phrases and sentences present which were not semantically related to the course text like “Any idea please?”, “Good day class” etc. This realization highlights some key challenges and points to a need to develop various approaches to automatically clean the text of real-world questions before performing the semantic search.

A look at some of the questions that our models got wrong provided interesting insights. The model’s misclassifications sometimes happened when the model retrieved an answer that could either partially or completely also answer the question but was not the designated correct answer. In the real world, these answers by our model will be sufficient to address the questions of students. Our current evaluation approach assumes that there is only one answer and also that only one paragraph is needed to answer a question. Hence, relaxing this assumption in future evaluations would improve the recognition results. Further investigation is also needed on the best way to combine partial answers in different paragraphs to provide one answer.

Computing top 3 and 5 accuracy is one step in that direction exploring multiple answers. The results comparing the top 1, 3 and 5 performance are shown in Table \ref{tab:top_n_results}. As expected, the accuracy results improved for top 3 and top 5 even getting up to 100\% with similar results for English and French questions. These results suggest that returning top 3 or 5 answers would enable students get the correct answers they need. In a course setting, returning 3 answers, for example, should not be overwhelming hence worth exploring especially if it could improve Kwame's real-world accuracy.

The closest work that we can compare our accuracy results to is the work of Zylich et al. \cite{zylich2020} that we described in the related work section. For their work, they attempted to correctly answer 18 factual physics questions which are similar to our coding content questions (20 quiz type and 12 student type). Their document retrieval approach which is analogous to our QA system had 44.4\%, 88.9\% and 88.9\% top 1, 3 and 5 accuracies respectively. Our best system had 58.3\% (58.3\%), 83.3\% (80\%) and 100\% (91.7\%) top 1, 3 and 5 accuracies for the student QA type for English (French). Hence, our results are overall slightly better than the work of Zylich et al. \cite{zylich2020} Another work to compare with is by Rana \cite{rana2019}. They implemented an approach for semantic search to answer questions that individuals asked about a university, based on information on the university’s website. One of their approaches is similar to ours and it involved document retrieval and paragraph selection using BERT. Their approach achieved 56\% accuracy. Our result of 58.3\% is also slightly better.

\section{Limitation and Future Work}
In this work, we did not evaluate Kwame in a real-world context of an online course. We will do this in the future and evaluate the real-world performance using a platform like Piazza which has been used extensively by Goel for real-word deployments \cite{goel2020}. The deployment would also entail Kwame returning figures as part of answers if the answer contains a reference to a figure in the lesson note. We will do the evaluation quantitatively (accuracy, upvotes or downvotes, response rate) and qualitatively (students’ feedback on Kwame).

Additionally, we used a small sample of QA pairs and sentences in this work — only the text from the first lesson. Future work will use all lessons and compare performance across lessons. Also, we will explore various approaches that could potentially increase the accuracy and reduce the time to retrieve the answers by using tags as filters (e.g. \#lesson1) to retrieve only the possible answers whose tag(s) match the tag of the question before performing comparison via cosine similarity.

Finally, the data and experiments in this work assume that all questions in the course are answerable by Kwame. Ideally, in a course, Kwame will not answer some questions if the confidence score is below a certain threshold. Further work is needed to detect and evaluate those cases.

\section{Conclusion}
In this work, we developed a bilingual AI TA — Kwame — to answer students’ questions from our online introductory coding course, SuaCode in English and French. Kwame is an SBERT-based question-answering (QA) system that we trained and evaluated using question-answer pairs created from the course’s quizzes and students’ questions in past cohorts. We compared the system to TF-IDF and Universal Sentence Encoder. Our results showed that SBERT performed the worst for duration (6 secs per question) but the best for accuracy, and fine-tuning on our course data improved the result. Also, returning the top 3 and 5 answers improved the results. Nonetheless, key challenges remain such as having a higher accuracy for real-world questions. 

Our long-term vision is to leverage the power of AI to democratize coding education across Africa using smartphones. Specifically, we aim to use AI to make learning more personalized for our students through providing accurate and timely answers to questions in our courses currently in English and French and in the future, in various African languages such as Twi (Ghana),  Yoruba (Nigeria), Swahili (Kenya, Tanzania), Zulu (South Africa). Kwame is an important part of achieving this vision by making it easy for students to get quick and accurate answers to questions in our SuaCode courses.

\section{Acknowledgement}
We are grateful to Professor Elloitt Ash for his mentoring and Victor Kumbol for helpful discussions.


\bibliographystyle{IEEEbib}
\bibliography{refs}

\begin{thebibliography}{10}

\bibitem{boateng2018}
George Boateng and Victor Kumbol,
\newblock ``Project iswest: Promoting a culture of innovation in africa through
  stem,''
\newblock in {\em 2018 IEEE Integrated STEM Education Conference (ISEC)}. IEEE,
  2018, pp. 104--111.

\bibitem{boateng2019}
George Boateng, Victor Wumbor-Apin Kumbol, and Prince~Steven Annor,
\newblock ``Keep calm and code on your phone: A pilot of suacode, an online
  smartphone-based coding course,''
\newblock in {\em Proceedings of the 8th Computer Science Education Research
  Conference}, 2019, pp. 9--14.

\bibitem{suacodeafrica}
``Nsesa runs suacode africa — africa’s first smartphone-based online coding
  course,'' https://nsesafoundation.org/nsesa-runs-suacode-africa/.

\bibitem{goel2016}
Ashok~K Goel and Lalith Polepeddi,
\newblock ``Jill watson: A virtual teaching assistant for online education,''
\newblock Tech. {R}ep., Georgia Institute of Technology, 2016.

\bibitem{goel2020}
Ashok Goel,
\newblock ``Ai-powered learning: Making education accessible, affordable, and
  achievable,''
\newblock {\em arXiv preprint arXiv:2006.01908}, 2020.

\bibitem{benedetto2019}
Luca Benedetto and Paolo Cremonesi,
\newblock ``Rexy, a configurable application for building virtual teaching
  assistants,''
\newblock in {\em IFIP Conference on Human-Computer Interaction}. Springer,
  2019, pp. 233--241.

\bibitem{zylich2020}
Brian Zylich, Adam Viola, Brokk Toggerson, Lara Al-Hariri, and Andrew Lan,
\newblock ``Exploring automated question answering methods for teaching
  assistance,''
\newblock in {\em International Conference on Artificial Intelligence in
  Education}. Springer, 2020, pp. 610--622.

\bibitem{suacode}
``Suacode - smartphone-based coding course,''
  https://github.com/Suacode-app/Suacode/blob/master/README.md.

\bibitem{suacodeafrica2}
``Suacode africa 2.0: Teaching coding online to africans using smartphones
  during covid-19,''
  https://www.c4dhi.org/news/lecture-by-boateng-suacode-africa-20210122/.

\bibitem{rajpurkar2016}
Pranav Rajpurkar, Jian Zhang, Konstantin Lopyrev, and Percy Liang,
\newblock ``Squad: 100,000+ questions for machine comprehension of text,''
\newblock {\em arXiv preprint arXiv:1606.05250}, 2016.

\bibitem{chen2017}
Danqi Chen, Adam Fisch, Jason Weston, and Antoine Bordes,
\newblock ``Reading wikipedia to answer open-domain questions,''
\newblock {\em arXiv preprint arXiv:1704.00051}, 2017.

\bibitem{devlin2018}
Jacob Devlin, Ming-Wei Chang, Kenton Lee, and Kristina Toutanova,
\newblock ``Bert: Pre-training of deep bidirectional transformers for language
  understanding,''
\newblock {\em arXiv preprint arXiv:1810.04805}, 2018.

\bibitem{yang2015}
Yi~Yang, Wen-tau Yih, and Christopher Meek,
\newblock ``Wikiqa: A challenge dataset for open-domain question answering,''
\newblock in {\em Proceedings of the 2015 conference on empirical methods in
  natural language processing}, 2015, pp. 2013--2018.

\bibitem{kwiatkowski2019}
Tom Kwiatkowski, Jennimaria Palomaki, Olivia Redfield, Michael Collins, Ankur
  Parikh, Chris Alberti, Danielle Epstein, Illia Polosukhin, Jacob Devlin,
  Kenton Lee, et~al.,
\newblock ``Natural questions: a benchmark for question answering research,''
\newblock {\em Transactions of the Association for Computational Linguistics},
  vol. 7, pp. 453--466, 2019.

\bibitem{garg2019}
Siddhant Garg, Thuy Vu, and Alessandro Moschitti,
\newblock ``Tanda: Transfer and adapt pre-trained transformer models for answer
  sentence selection,''
\newblock {\em arXiv preprint arXiv:1911.04118}, 2019.

\bibitem{reimers2019}
Nils Reimers and Iryna Gurevych,
\newblock ``Sentence-bert: Sentence embeddings using siamese bert-networks,''
\newblock {\em arXiv preprint arXiv:1908.10084}, 2019.

\bibitem{reimers2020}
Nils Reimers and Iryna Gurevych,
\newblock ``Making monolingual sentence embeddings multilingual using knowledge
  distillation,''
\newblock in {\em Proceedings of the 2020 Conference on Empirical Methods in
  Natural Language Processing}. 11 2020, Association for Computational
  Linguistics.

\bibitem{rana2019}
Muhammad Rana,
\newblock ``Eaglebot: A chatbot based multi-tier question answering system for
  retrieving answers from heterogeneous sources using bert,''
\newblock 2019.

\end{thebibliography}

\end{document}